\documentclass{article}

\pdfoutput=1 

\usepackage{arxiv}

\usepackage[utf8]{inputenc} 
\usepackage[T1]{fontenc}    
\usepackage{url}            
\usepackage{booktabs}       
\usepackage{amsfonts}       
\usepackage{nicefrac}       
\usepackage{microtype}      
\usepackage{lipsum}		
\usepackage{graphicx}
\usepackage{doi}
\usepackage{xcolor}
\usepackage[square,numbers]{natbib}

\usepackage{times}
\usepackage{epsfig}
\usepackage{amsmath}
\usepackage{amssymb}
\usepackage{dirtytalk}
\usepackage{multirow}
\usepackage{subfig}

\begin{document}

\title{Few-Shot Batch Incremental Road Object Detection via Detector Fusion}

\author{Anuj Tambwekar \thanks{Work done as an intern at Intel Corporation}\\
PES University\\
{\tt\small anujstam@gmail.com}

\and
Kshitij Agrawal \qquad
 Anay Majee \qquad
 Anbumani Subramanian\\
Intel Corporation\\
{\tt\small firstname.lastname@intel.com}


}

\maketitle

\definecolor{orange}{RGB}{246, 178, 107}
\definecolor{archblue}{RGB}{164, 194, 244}
\definecolor{archgreen}{RGB}{147, 196, 125}

\begin{abstract}
Incremental few-shot learning has emerged as a new and challenging area in deep learning, whose objective is to train deep learning models using very few samples of new class data, and none of the old class data. In this work we tackle the problem of batch incremental few-shot road object detection using data from the India Driving Dataset (IDD). Our approach, DualFusion, combines object detectors in a manner that allows us to learn to detect rare objects with very limited data, all without severely degrading the performance of the detector on the abundant classes. In the IDD OpenSet incremental few-shot detection task, we achieve a $mAP_{50}$ score of 40.0 on the base classes and an overall $mAP_{50}$ score of 38.8, both of which are the highest to date. In the COCO batch incremental few-shot detection task, we achieve a novel $AP$ score of 9.9, surpassing the state-of-the-art novel class performance on the same by over 6.6 times.
\end{abstract}

\section{Introduction}

Object detection is a challenging task that aims to detect the presence and location of desired objects in an image. It is a crucial part of smart transportation, where it is used by both autonomous vehicles, as well as advanced driver assistant systems (ADAS) \cite{10.3389/frobt.2015.00029}. 

Current state of the art object detection networks such as Faster R-CNN \cite{faster-rcnn}, YOLO \cite{yolo} and RetinaNet \cite{retinanet} perform well on canonical object detection datasets such as PASCAL VOC \cite{pascal_voc} and COCO \cite{ms-coco}. These object detection networks are reliant on large-scale annotated datasets for training. Acquiring such large volumes of data is both expensive or requires tremendous amount of human effort. The traditional object detection datasets, are well-balanced and contain a sufficiently large number of examples per category. However, in a real world, classes of interest may be rare and hard to come by, thus resulting in a data scarcity due to the nature of the objects in question. This problem is one of the key challenges in detecting rare objects found on Indian roads \cite{majee2021fewshot}, where there is a significant class imbalance between the frequently occurring classes, and the rarer classes. 

Few shot learning has been proposed to overcome these challenges and learn with limited amount of data. While significant progress has been made in this area in both the classification \cite{fewshot_vinyals, proto_fewshot, sung2018learning} and detection \cite{FSOD_KT, feat_reweight, fsdet, fewx, meta_rcnn} tasks, most techniques assume that the data from the abundant classes is always available. This requirement cannot always be met, as there may be scenarios where we require to increase the capability of the model to detect new classes without access to old data. For example if the models are deployed in autonomous vehicles and the vehicle needs to detect new objects on the road. This challenge of learning to detect new classes in addition to existing classes, while only being exposed to the new classes, is termed incremental object detection. As described in \cite{ONCE}, learning one class at a time is termed class incremental object detection, while learning to detect multiple new classes in one attempt is termed batch incremental object detection. Few-shot object detection techniques also suffer from class confusion \cite{metadet}, a phenomenon where objects are located correctly, but classified into the wrong class. Base-novel confusion refers to this incorrect classification, where an object belonging to a base class is incorrectly labeled as an object of a novel class, or vice-versa.  

Most existing works treat few-shot detection and incremental detection as two separate tasks, when in reality it is important to address both simultaneously. One of the key requirements of existing incremental learning techniques \cite{ShmelkovIncremental, chen_incremental_distillation, hao_architecture_incremental} is a reliance on having a large amount of data corresponding to the new classes, making them ineffective in a few-shot setting. Additionally, incremental learning techniques are prone to \textit{catastrophic forgetting} \cite{french1999catastrophic, mccloskey1989catastrophic}, a phenomenon where the incremental detector fails to detect the previously learnt classes and thus forgets the previously learnt features. In a safety-critical environment such as an ADAS or an autonomous vehicle, it is crucial to ensure that none of the the prior learning is forgotten, making existing techniques unsuitable for these tasks.

In this work, we propose \emph{DualFusion}, a method to address few-shot batch incremental learning to detect road objects using only 10 annotated instances of each novel class, and without any base class data. The proposed architecture combines class-specific object detectors using a proposal segregation technique. This improves the classification ability of the combined detector to detect novel and base classes from overlapping regions of interest, while still retaining knowledge from the individual classifiers as no re-training is involved. We specifically design this technique for contextual datasets like road object detection where rare novel classes are found alongside bases classes.

The rest of this paper is divided as follows - Section \ref{sec:background} provides an overview of the background and related work. Section \ref{sec:df} describes the architecture of DualFusion in detail, along with its training strategy. Section \ref{sec:experiments} goes over the experimental setup used to validate DualFusion on road object detection with the India Driving Dataset (IDD)\cite{IDD} as well as more generalized cross-domain setting on COCO, where it achieves SOTA performance on novel classes. Finally in Section \ref{sec:conclusion} we conclude and discuss the advantages, drawbacks and scope for extension.

\section{Background and Related Work} \label{sec:background}

\subsection{Object Detection}
Standard object detection techniques require large amount of data to train and aim to localize and classify objects present in an image. They are broadly categorized into two groups - single stage detectors and two-stage detectors. Single Stage detectors such as SSD \cite{SSD}, Retinanet \cite{retinanet} and YOLO \cite{yolo} extract regions of interest and classify objects in a single step. Meanwhile, two stage detectors such as Faster R-CNN \cite{faster-rcnn} first identify regions of interest via a Region Proposal Network (RPN), and then determine the bounds of the object and its class based on the contents of the region of interest. These techniques cannot be applied directly when dealing with heavy class imbalance and when only a limited amount of data is available as shown by prior work such as \cite{feat_reweight, fsdet}.

\subsection{Few-Shot Object Detection}
Few-shot learning \cite{few_shot_survey} is a technique that learns a data distribution using very limited training data. Formally, if $N$ new classes are learnt in the few-shot setting, and there are $K$ training examples per class, it is termed $N$-way $K$-shot learning.

Recent successful few-shot object detection techniques are episodic approaches that operate by reweighting the features between classes. such as Feature-Reweight \cite{feat_reweight}. Meta-RCNN \cite{meta_rcnn} extends the previous approach to a Faster R-CNN based network. It is a meta-learning technique that uses an episodic training regime to attempt to train the network to learn how to learn better. In a similar fashion Kim et al. \cite{FSOD_KT} uses a meta-learner that employs knowledge transfer to create class-vector prototypes that are used to provide weights to the features. MetaDet \cite{metadet}, uses meta-learning to create a model that learns to transform weights learnt during the few-shot training stage to large-scale training weights.

Two-stage finetuning \cite{fsdet}, provided a different viewpoint, and instead suggested that the lower layers of the detectors produced class agnostic features, while the higher layers produced class specific features. The technique, known as FsDet simply called for fine-tuning the final layer using cosine similarity in order to learn the new classes. However, FsDet is reliant on a well balanced training dataset which contains both the base and the novel classes in the fine-tuning stage. This reliance on base class labels during the fine tuning stage makes it unsuitable for an incremental learning setting.

 One of the best performing few-shot object detection techniques is FewX \cite{fewx}, which uses attention maps as a way to search for the presence of the novel classes in an image. In theory, this allows the network to detect novel classes without any kind of fine-tuning, if trained on a sufficiently varied dataset, but for context specific situations, fine-tuning greatly improves performance. While this architecture claims the highest performance on the COCO \cite{ms-coco} dataset, it suffers from the glaring limitation of only being able to detect the novel classes. FewX uses attention as a mechanism to find candidate proposals and then classify objects in an image. Hence, if no base classes are present in the support set, FewX will never be able to detect them. This makes it an impractical solution when any of the classes have sufficient data and need not be treated as few-shot, as few-shot performance is always less than standard training performance. It also poses a challenge when access to the base class data is withheld.

\begin{figure*}[tbp]
    \centering
    \includegraphics[width=\textwidth]{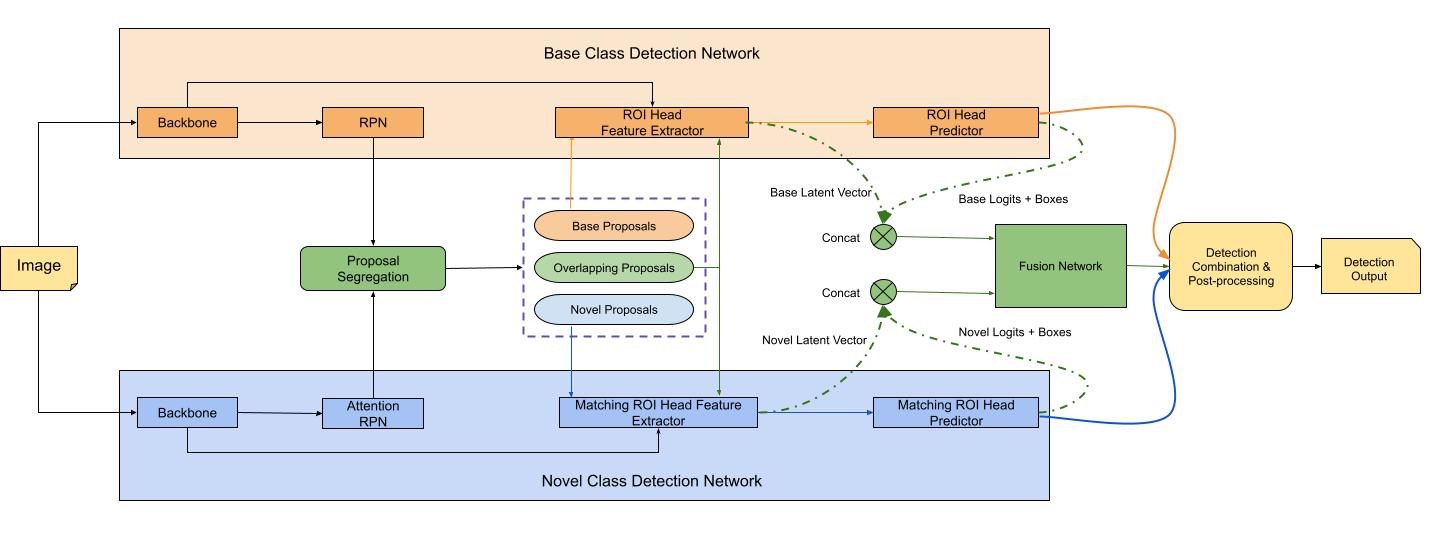}
    \caption{Overall architecture of \emph{DualFusion}. Components in \textcolor{orange}{orange} denote parts of the Faster-RCNN base detector, those in \textcolor{archblue}{blue} denote parts of the FewX based novel detector, and those in \textcolor{archgreen}{green} denote the Fusion Network. Coloured arrows denote the flow of the various proposal categories - orange for base categories, blue for novel, and green for overlapping.}
    \label{fig:dualfusion_arch}
\end{figure*}

\subsection{Incremental Object Detection}
Incremental object detection learns to detect new classes in addition to the existing classes. In this setting, the detectors are trained to retain the base class performance and while only being exposed to the new classes. Incremental learning can be divided into two categories based on the number of classes being added in every stage - single class incremental learning where new categories are introduced one at a time, and batch incremental learning, where multiple new categories are learnt together.

Most existing incremental object detection techniques \cite{Shmelkov2017IncrementalLO, chen_incremental_distillation, hao_architecture_incremental} operate on knowledge distillation \cite{hinton_distil}. \cite{Shmelkov2017IncrementalLO} uses a balanced loss combing distillation as well as cross-entropy to teach a new network to detect new classes, as well as to attempt to transfer the knowledge from the old detector. \cite{chen_incremental_distillation} uses intermediate proposals, along with two terms they call hint loss and confidence loss, to transfer feature maps from the original model. \cite{hao_architecture_incremental} expands the foreground domain of the region proposal network, and uses a nearest-prototype classifier as well as distillation loss to expand the detector's ability to add in new classes. The chief drawback with all these techniques is that they assume an abundance of data corresponding to the new classes. The current state-of-the-art technique, Open-ended Centre Net (ONCE) \cite{ONCE}, is the only existing few-shot incremental object detection network, and works by re-weighting image features based on a class code generator. The reported quantitative results for the novel classes are, however, very low.

\section{Our Approach : DualFusion} \label{sec:df}

In this work we target few-shot batch incremental object detection. In this setting, the objective is to learn to detect multiple novel classes at once, while retaining the ability to detect the base classes, using only novel class data in the incremental addition stage. 

One of the biggest challenges with any incremental learning scenario is \textit{catastrophic forgetting} \cite{french1999catastrophic}. While techniques such as knowledge distillation \cite{hinton_distil} have been proposed to combat this limitation, they still result in a heavy loss of performance on the base classes \cite{etoe_incremental,Shmelkov2017IncrementalLO}. We propose a simple yet powerful solution to circumvent this problem - instead of re-training the existing network, or training a new network using a distillation loss, we retain the original network, and combine it with another network specifically trained to discriminate for novel classes. Our approach can thus be extended to detect multiple novel classes incrementally without the need for base class data.

Formally, we can define the task as follows - Given a set of base classes $B$, a detector capable of detecting the base classes $D_B$, and novel classes $N$, our goal is to create a new detector $D'$, capable of detecting all classes from $B$ and $N$, while only having access to 10 annotated instances of each class in $N$, and no annotated instances of classes in $B$.

\subsection{Architecture}
The DualFusion architecture comprises of 3 components
\begin{itemize}
    \item A Faster R-CNN based base class detector to detect the base classes, that is trained only on base class data, with no exposure to novel classes.
    \item A few shot novel class detector based on FewX modified, so as to combine the outputs of each attention head into an intermediary layer before the classifier and box regressor. 
    \item The Fusion Network. This network is designed to classify objects that are flagged as possible candidates by both the base class and the novel class detector. It is akin to an ROI head predictor that learns to combine the features obtained by the base and novel detector to classify objects.
\end{itemize}
Figure \ref{fig:dualfusion_arch} provides a pictorial representation of the architecture. The black arrows indicate the flow of data before and after the proposal segregation step. The coloured arrows indicate the flow of data corresponding to the various proposal categories. Orange arrows denote the flow of data corresponding to the base class proposals, Blue arrows denote the flow of data corresponding to the flow of novel class proposals, and green arrows denote the flow of data corresponding to the overlapping proposals. 

\subsubsection{Architectural Setup and Flow}
The final DualFusion model comprises of the three models mentioned earlier. Details about the training process are described subsequently in Section \ref{sec:training}. In short, the faster RCNN base class detector is trained only on base classes, the novel class FewX detector is only trained on few-shot samples of the novel classes, and the Fusion network is trained using features extracted from the mined dataset whose collection is described in the Training section.

An input image is passed through the backbone and RPN of both the novel class and base class detectors. The proposals are then segregated into 3 categories - base proposal regions, novel proposal regions and overlapping proposal regions. We describe why this is needed and how it is exactly done in Section \ref{sec:proposalsegregation}. The base proposals are passed through the base class detector as usual to obtain a set of features, which are then converted into a detection output. In a similar fashion, the novel class detector acts on novel proposals and identifies the rare novel classes.

The overlapping proposals are passed to \textbf{both} detectors' feature extraction networks, and the penultimate outputs and the predicted boxes and logits are extracted. These latent feature and prediction combinations are first passed through a single fully-connected layer, following which they are concatenated and passed to the fusion network, which learns to distinguish between the base and novel detections, thereby reducing base-novel confusion. Each input is passed to a corresponding two layer deep fully connected network that uses the SeLU \cite{selu} activation function - this allows the features to be transformed into the same feature space. The transformed features are then combined and passed through two layers with ReLU activation, for the class score and bounding boxes predictions. 

As result, we have a collection of three detection outputs at the end of a forward pass of an image. These detection outputs are combined, so as to remove overlapping predictions of the same region, and to reduce intra-novel confusion, by comparing the IoU between the three detection sets and selecting the detections with larger confidence scores whenever the IoU between detections from different detectors exceeds 0.5. This step is similar to the standard Non-Maximal Suppression (NMS) post-processing step of object detectors, however it considers the output of different detectors in the same region, as opposed to detections of the same class.

\subsubsection{Proposal Segregation} \label{sec:proposalsegregation}

The obvious question to ask is - \say{Why can't we just use two detectors?}. A simple strategy of using two different detectors does not work as novel objects with similar features may be incorrectly classified as base class objects and vice-versa. Additionally, regions with features present in both the base and novel classes may be detected by both detectors, and thus the same region may be marked as two different objects (see Figure \ref{fig:fusion_cmp} for an example of the same). We term this phenomenon as base-novel confusion. Our design of fusing the intermediate outputs of each of the networks allows for a degree of communication between detectors, and thereby reduces this confusion. We utilize FewX as a novel class detector as it is, at the time of writing, the best performing few-shot detection model on COCO. Additionally, it is a two-stage approach, which allows us to implement proposal segregation.

Proposal segregation allows us to retain base and novel class performance by using the appropriate detector in regions without confusion. We first generate a collection of candidate proposals from the base detector $P_{B}$ and candidate proposals from the novel detector $P_{N}$. Following this, we compute the element-wise Intersection over Area (IoA) between $P_{B}$ and $P_{N}$. The proposals from $P_{B}$ whose IoA with every proposal from $P_{N}$ is below a certain threshold are considered valid base proposals. Likewise all proposals from $P_{N}$ whose IoA with every proposal from $P_{B}$ is below the threshold are considered valid novel proposals. The rest of the proposals are considered as overlapping proposals. We use IoA instead of Intersection over Union (IoU), as it provides a better estimate of proposal overlap between base and novel classes, and used a threshold value of 0.5. 

\subsection{Training}\label{sec:training}
DualFusion is trained over a series of 4 steps
\begin{itemize}
    \item \textbf{Step 1. Base Model Training}: To setup the base trained models, we train a Faster R-CNN based detector to detect IDD base classes. Following this, we train a FewX based model on the IDD base classes, using just 10 images from each class - this is done in order to allow for domain adaptation. The Faster R-CNN detector is frozen and from this point forward, we assume we have no access to the base training data.
    \item \textbf{Step 2. Novel Model Fine-Tuning} : We fine-tune the FewX based few-shot object detector specifically to detect novel classes. The only data we have access to at this point are driving scene images where novel classes are present. We assume 10 images per novel class, giving us a total of 10 $\times$ N training samples, where N is the number of novel categories. Furthermore, we only use the ground truth annotations of the novel classes present in these training samples.
    \item \textbf{Step 3. Fusion Network Dataset Creation} : Due to the nature of images in IDD, the novel classes occur in driving scenes which contain the presence of base classes. As such, while the training samples only contain ground truth annotations of novel classes, there exist un-annotated instances of some of the base class objects in the training data. By using the predictions from base class detector, we generate pseudo-ground truth labels for the base classes and use them for training. Any objects detected with a bounding box IoU of over 50\% with a novel class are removed. It is not required, nor expected of the data to contain instances of every base object class as part of this scenario, as the goal is to simply teach the detector how to distinguish between the features of the base and novel detectors.
    \item \textbf{Step 4. Fusion Network Training} : Combining the pseudo labels of base class from Step 3, we generate the final training set. We train the fusion network that combines features from both the base and novel detectors to provide predictions in regions of high base-novel confusion.
\end{itemize}

\begin{figure}[htbp]
\begin{center}
\begin{tabular}{cc}
\subfloat[Tractor]{\includegraphics[width=0.35\linewidth, height = 0.3\linewidth]{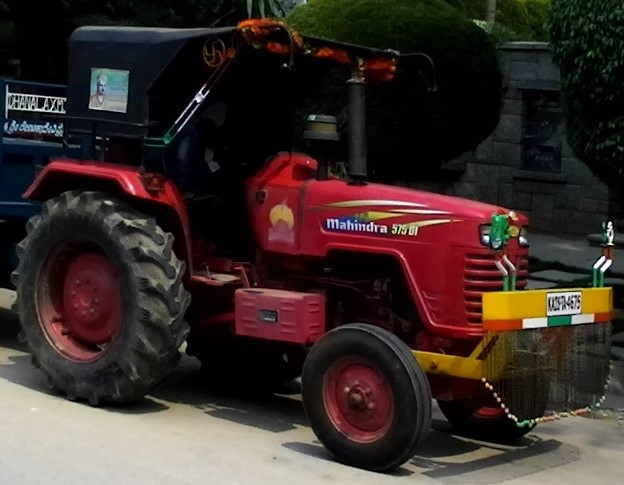}} &
\subfloat[Street Cart]{\includegraphics[width=0.35\linewidth, height = 0.3\linewidth]{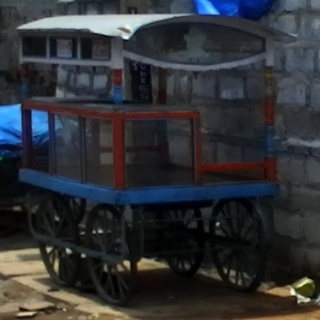}} \\  
\subfloat[Water Tanker]{\includegraphics[width=0.35\linewidth, height = 0.3\linewidth]{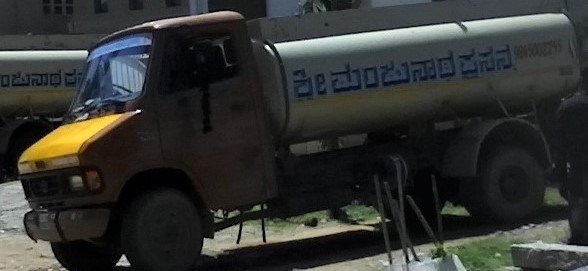}} &
\subfloat[Excavator]{\includegraphics[width=0.35\linewidth, height = 0.3\linewidth]{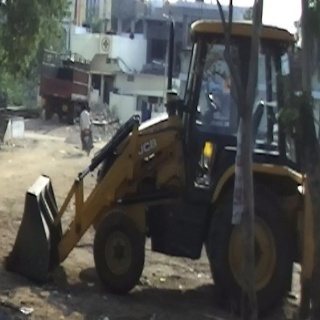}}\\
\end{tabular}
\end{center}
\caption{Examples of novel classes in IDD-OS}
\label{fig:novel_cls}
\end{figure}

\begin{figure*}[htbp]
\begin{center}
\begin{tabular}{cccc}

{\includegraphics[width = 0.22\textwidth]{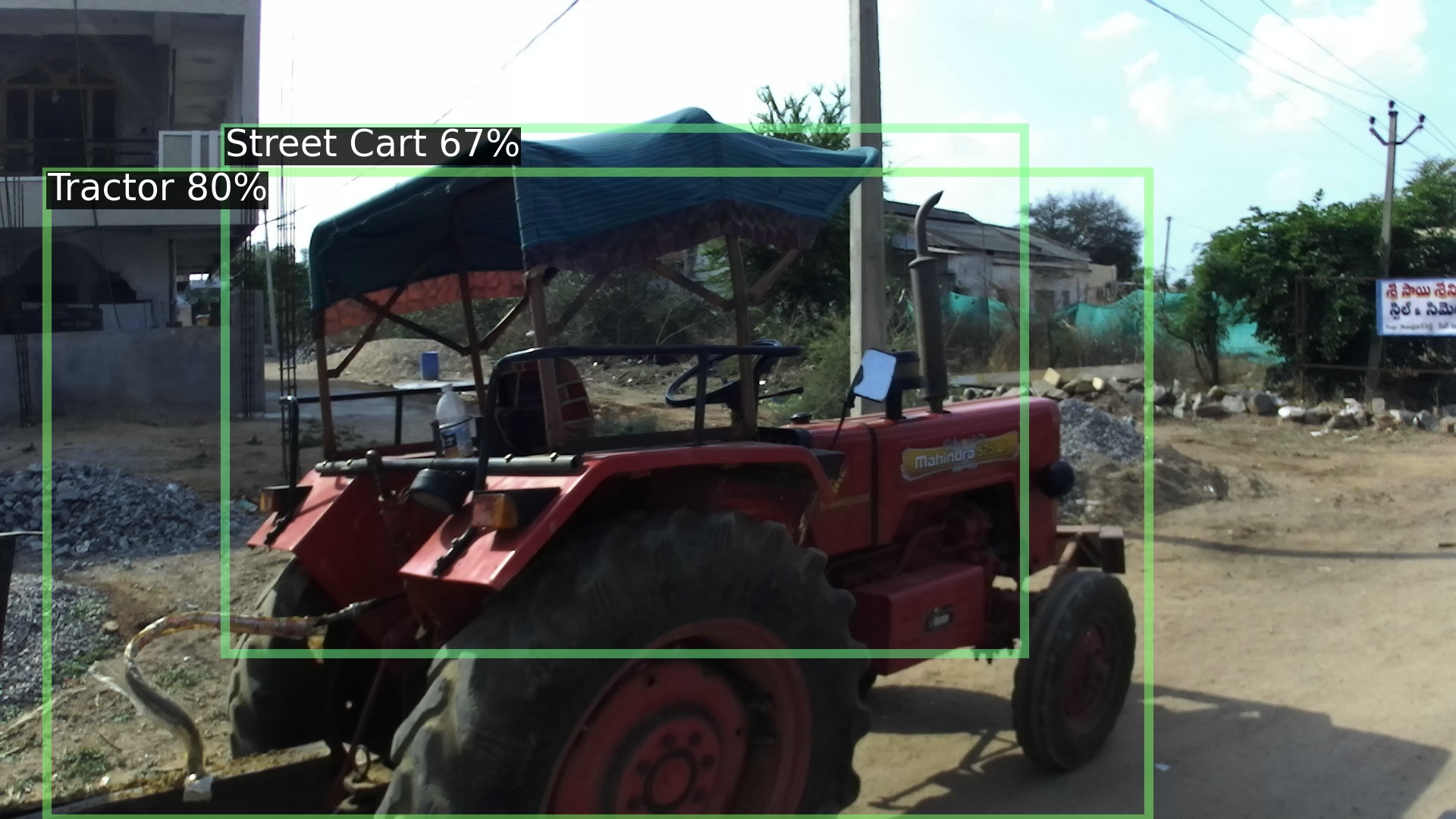}} &
{\includegraphics[width = 0.22\textwidth]{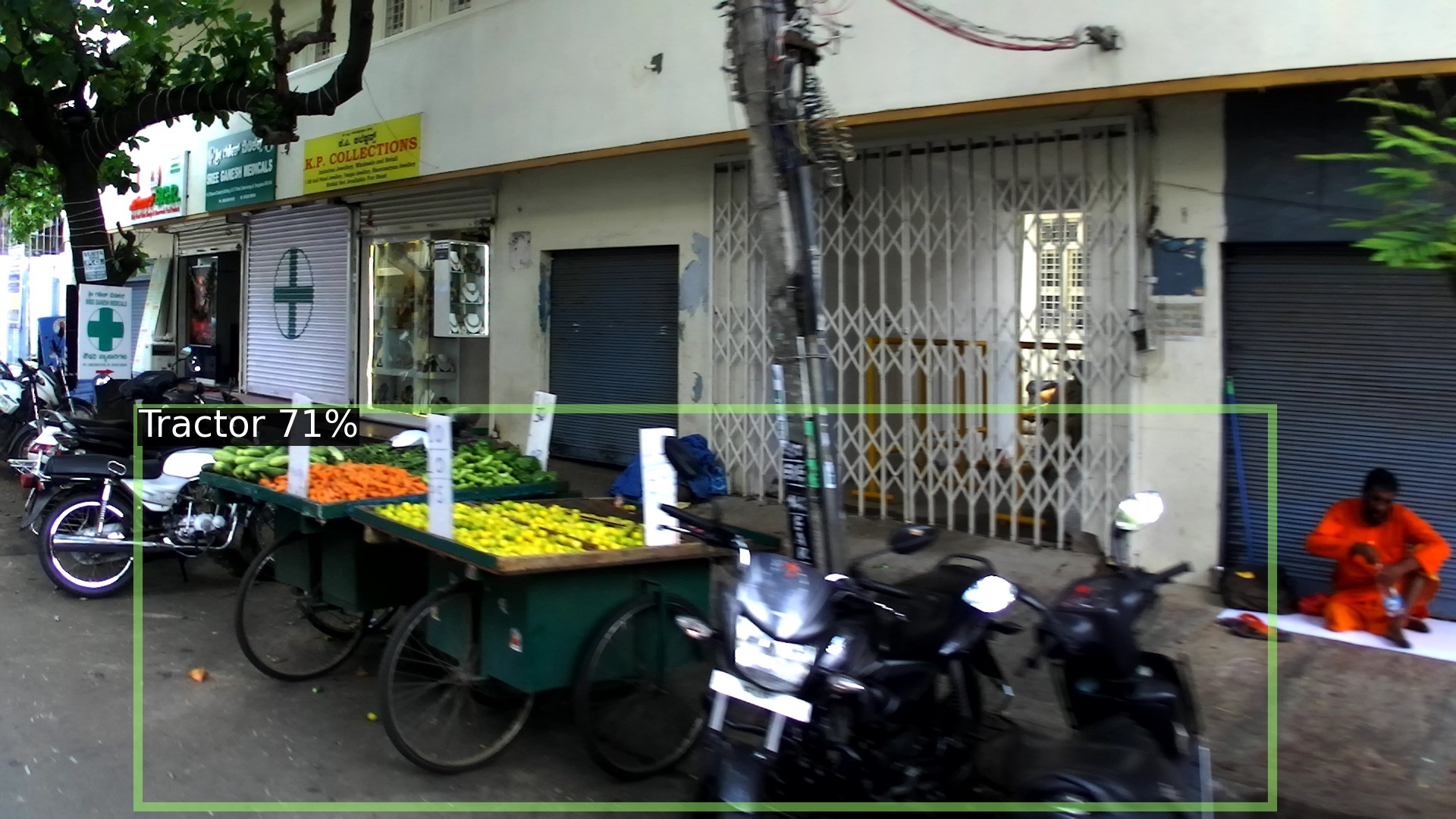}} &
{\includegraphics[width = 0.22\textwidth]{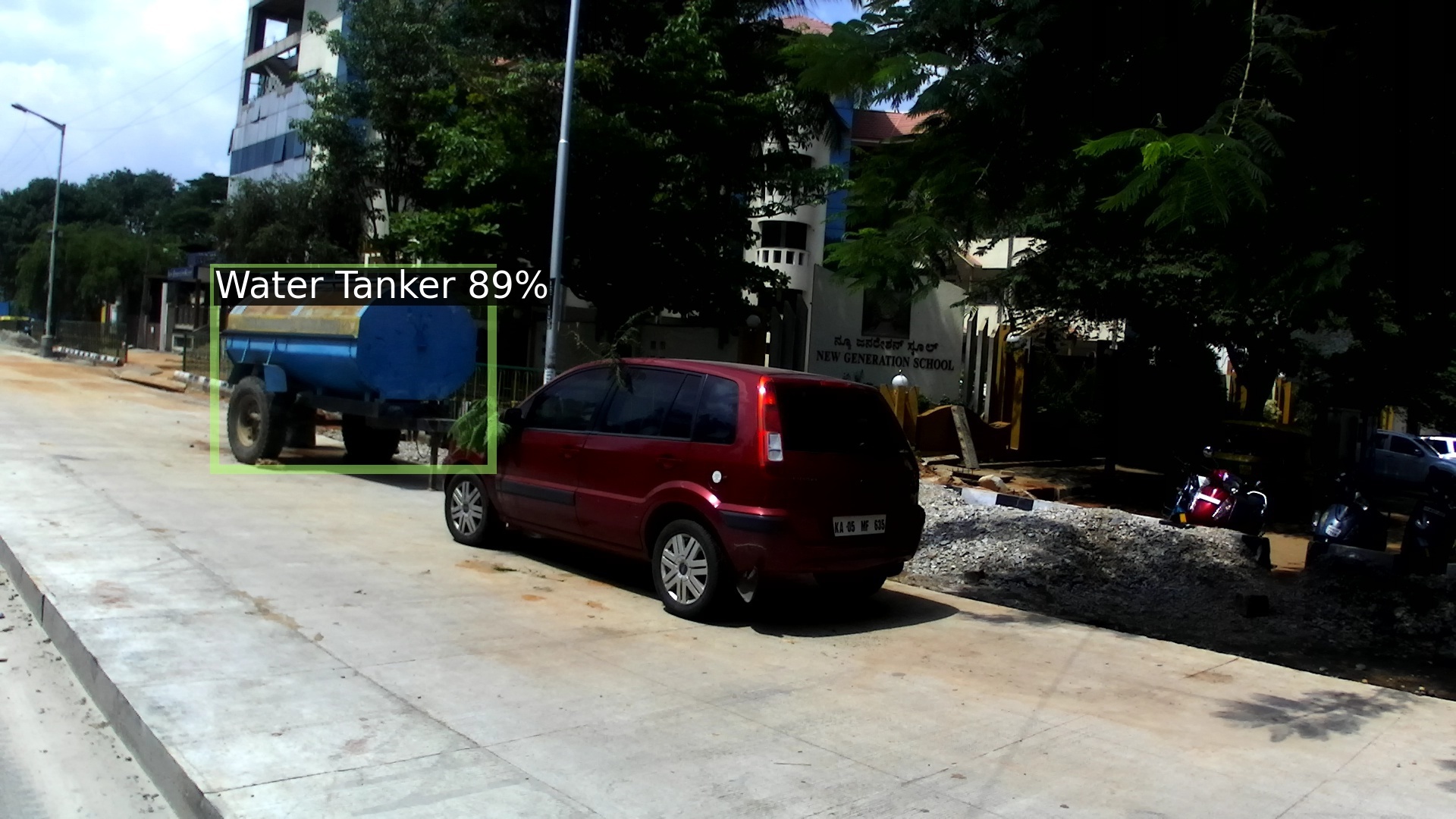}} &
{\includegraphics[width = 0.22\textwidth]{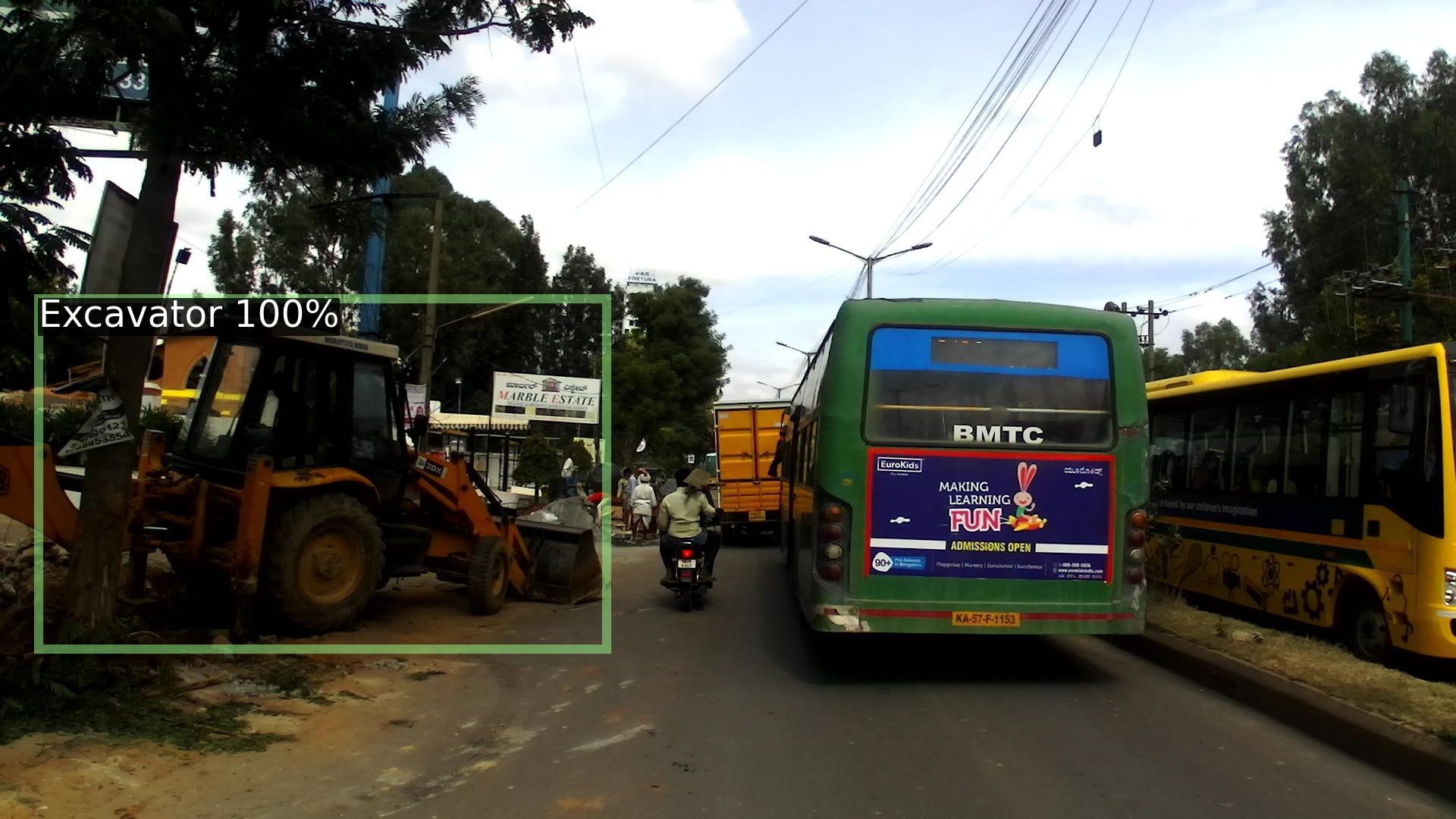}}\\

\subfloat[Tractor]{\includegraphics[width = 0.22\textwidth]{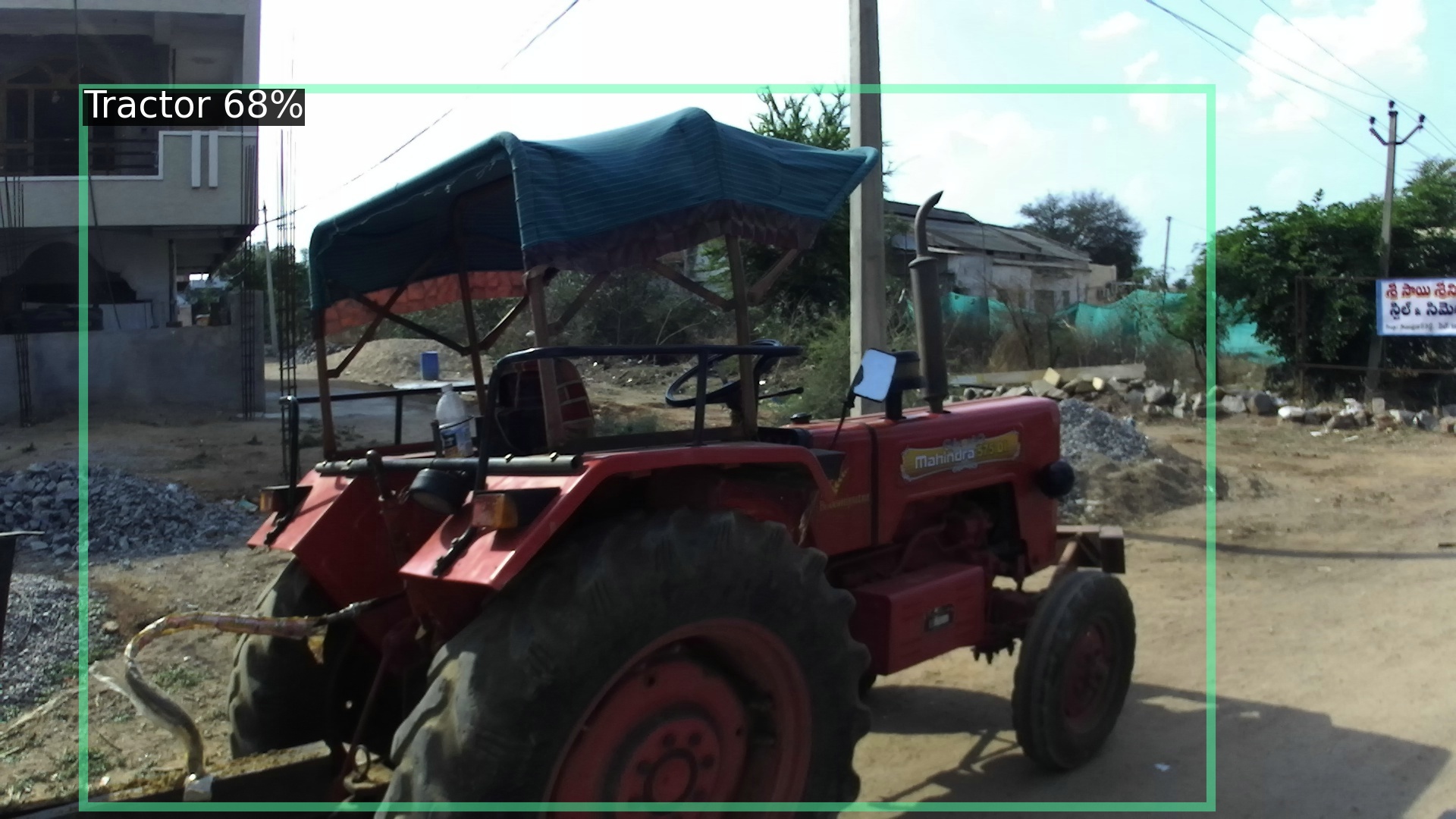}} &
\subfloat[Street Cart]{\includegraphics[width = 0.22\textwidth]{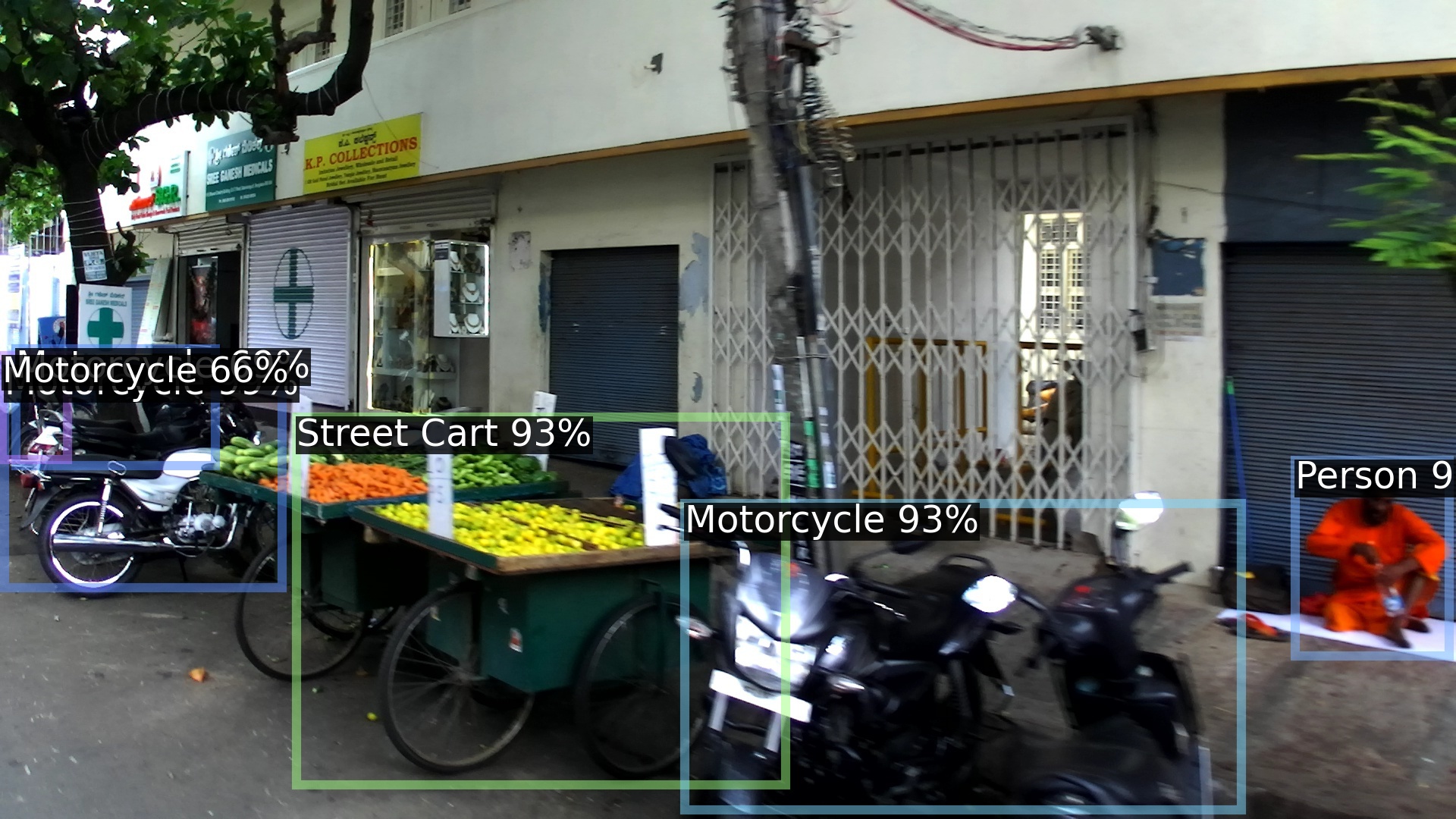}} &
\subfloat[Water Tanker]{\includegraphics[width = 0.22\textwidth]{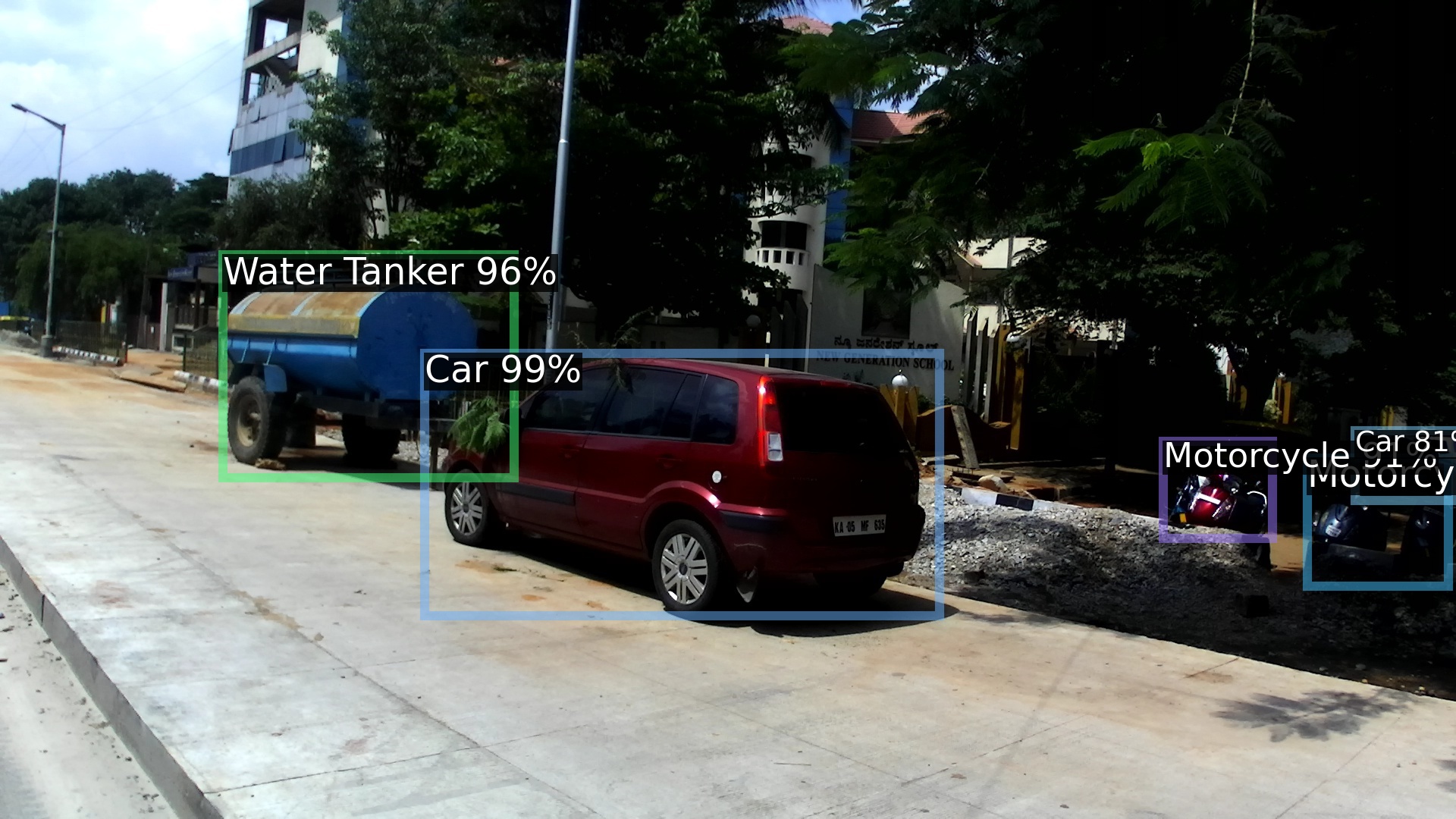}} &
\subfloat[Excavator]{\includegraphics[width = 0.22\textwidth]{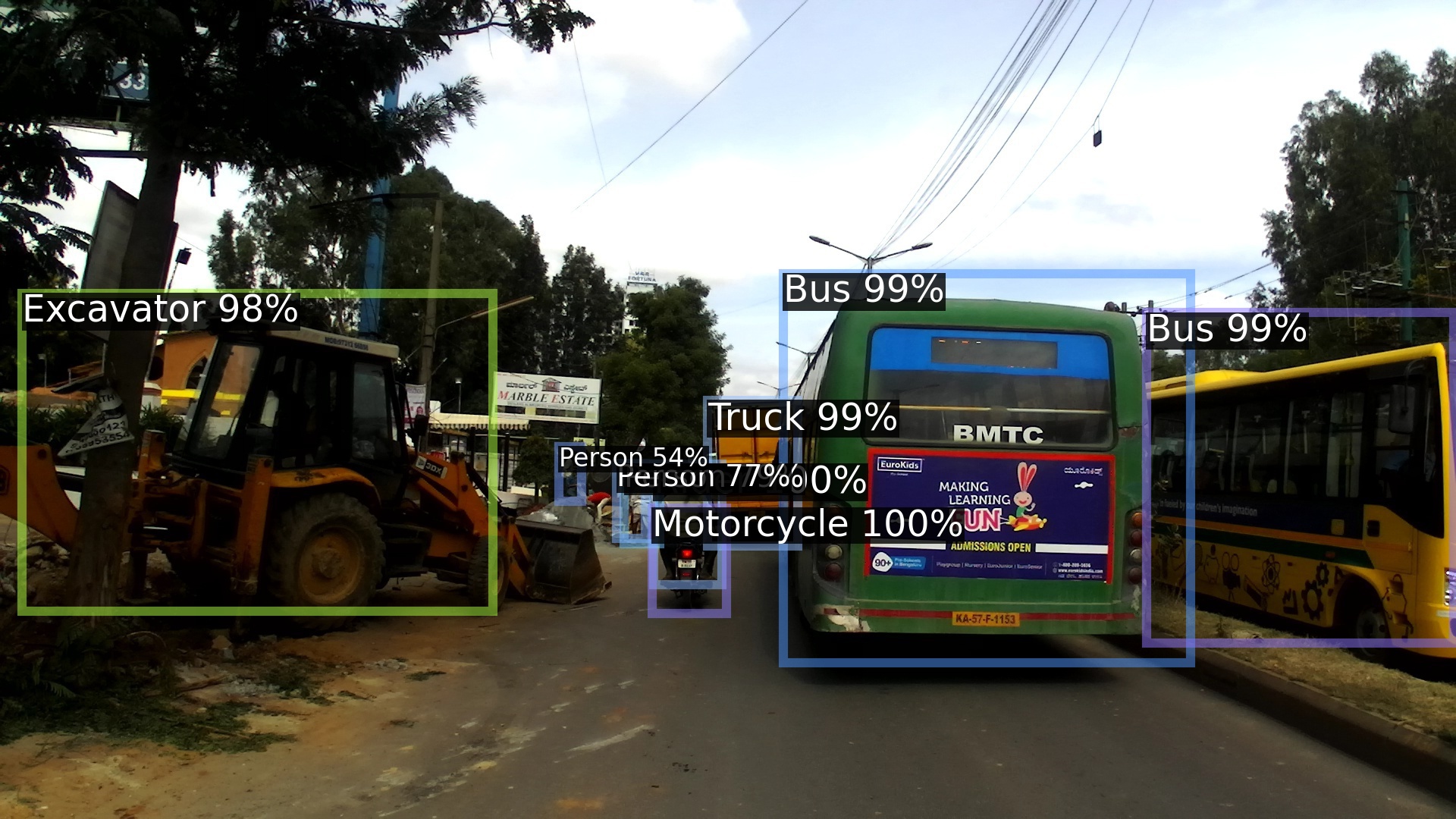}}\\

\end{tabular}
    
\end{center}
\caption{Examples of novel (green bounding boxes) and base (blue bounding boxes) class detections on IDD-OS via Faster R-CNN Finetuning (top row) and DualFusion (bottom row). Base class forgetting and erroneous novel class detections are clearly seen in Finetuning, but both are avoided by DualFusion.}
\label{fig:dualfusion_egs}
\end{figure*}

\section{Experimental Setup and Results} \label{sec:experiments}
To validate the efficacy of DualFusion as a batch incremental few-shot learning paradigm, we run a set of tests comparing it to existing techniques for both the India Driving Dataset (IDD) and COCO. 

\subsection{IDD}
For our experiments we use the IDD-OpenSet (IDD-OS) split as described by \cite{majee2021fewshot} over conventional driving datasets such as Berkeley Deepdrive \cite{bdd100k} and KITTI \cite{KITTI}, due to the fact that it possesses the additional challenge of large amount of class imbalance, along with a well-defined splits for consideration of base classes and novel classes. Examples of the objects belonging to the novel categories can be seen in Figure \ref{fig:novel_cls}.

Given an object detector $D$ trained on the IDD Base Set classes - \textit{Autorickshaw, Bicycle, Bus, Car, Person, Motorcycle, Rider, Traffic Light, Traffic Sign,} and \textit{Truck}, train a detector $D'$, that can detect the rare novel classes from IDD-OS - \textit{Tractor, Street Cart, Water Tanker} and \textit{Excavator}, in a batch incremental setting. The detector $D'$ is trained to detect novel classes with only 10 annotations of each, while simultaneously having access to the original detector $D$. 

We compare the performance of DualFusion to various techniques. We test two Faster R-CNN finetuning strategies. The first is standard Faster R-CNN Finetuning, where we simply add 4 additional outputs to the detector and train it on the novel classes. The second is Faster R-CNN Last-Layer Finetuning, where we finetune only the last layer. For both, we train for 2000 iterations at a learning rate of 0.002 and a learning rate decay step of 1000 iterations.
We also test FsDet without the rebalancing finetuning stage as discussed by the authors in \cite{fsdet}. In this step, the merged network is finetuned on a balanced dataset containg annotations from all the classes. As this requires base class data, we skip this step, and instead use the model obtained after combining the model trained on the base data, and the model obtained by training on novel data. Lastly, we compare with FewX. All five networks are initially trained on IDD-Base, following which, they are trained on only the 10 novel examples from IDD-OS. As per \cite{majee2021fewshot}, we report the $mAP_{50}$ score for each category. Feature Reweight \cite{feat_reweight} was not used due to out-of-memory issues while training. Our Faster R-CNN base detector roofline gave us a base class $mAP_{50}$ detection score of 50.1. We found that fine-tuning FewX was required in order to allow it to adapt to the driving domain. The FewX network and the DualFusion novel arm were trained for 2500 iterations, at a learning rate of 0.002 with steps at 1000 and 1500 iterations and a batch size of 2.

\textbf{Results} :
Table \ref{tab:dualfusion_cmp_idd_novel} shows a comparison between the $mAP_{50}$ scores of DualFusion and the prior mentioned Few-shot detection techniques. From the table we can observe that DualFusion is able to learn the novel classes extremely well, while significantly retaining base class knowledge, resulting in the highest base $mAP_{50}$ and the the highest overall $mAP_{50}$ performance. Additionally, DualFusion also performs better for certain novel classes (street cart and excavator) compared to FewX, due to its ability to reduce base-novel confusion. Our technique is able to obtain 80\% of the base class performance of a conventional Faster R-CNN, achieving a $mAP_{50}$ of 40.0, all while extending the ability of the detector to detect the novel classes. A few example detections predicted by DualFusion on IDD-OS can be seen in Figure \ref{fig:dualfusion_egs} where they are contrasted with detections predicted by Faster R-CNN finetuning method. As seen in the figure, while finetuning is able to learn to pick out novel classes, it depicts severe base class forgetting. DualFusion is able to retain base classes as well as provide better detections for the novel classes. While FewX does outperform DualFusion in two novel classes, it is unable to detect any base classes due to the absence of base class supports in the incremental setting, showcasing DualFusion's superiority.  

\begin{table*}[hbp]
\centering
\resizebox{\textwidth}{!}{  
\begin{tabular}{l|c|c|c|c|c|c|c}
\hline
\multicolumn{1}{c|}{\multirow{2}{*}{Method}} & 
\multicolumn{4}{c|}{Novel Classes ($mAP_{50}$)} & 
\multicolumn{1}{c|}{\multirow{2}{*}{$mAP_{50, novel}$}} & 
\multicolumn{1}{c|}{\multirow{2}{*}{$mAP_{50, base}$}} &
\multicolumn{1}{c}{\multirow{2}{*}{$mAP_{50, all}$}}\\
\cline{2-5}
\multicolumn{1}{c|}{}& Tractor & Street Cart & Water Tanker & Excavator & & &\\
\hline
Faster R-CNN Finetuning & 16.8  & 8.5  & 0.1  & 21.6  & 11.8  & 0.0 & 3.4  \\ 
Faster R-CNN Finetuning (only last layer) & 12.9  & 9.1  & 0.1  & 13.2  & 8.8  & 0.0 & 2.5  \\ 
FsDet (no rebalancing finetuning) & 0.0 &  0.0 & 0.0 & 0.0 & 0.0 & 9.42 & 6.73 \\ 
FewX                & \textbf{48.9} & \textbf{50.1} & 31.5 & 22.0 & \textbf{38.1} &  -  & 10.9 \\ \hline
\textbf{DualFusion (ours)}   & 30.5 & 28.6 & \textbf{32.0} & \textbf{42.1} & 33.3 &  \textbf{40.0}  & \textbf{38.8} \\ \hline
\end{tabular}%
}
\caption{$mAP_{50}$ Scores on the IDD Open Set 10-shot Batch Incremental Task. Cells filled with '-' indicate that the technique cannot be used for the task, and hence a score is not applicable. While FewX performs well on novel classes, it cannot detect the base classes. On the other hand, DualFusion achieves the highest base class and overall class performance on the task, and is clearly able to retain base class knowledge while learning to detect the novel classes.}
\label{tab:dualfusion_cmp_idd_novel}
\end{table*}

\begin{figure*}[bp]
\begin{center}
\begin{tabular}{cc}
\subfloat[Without FusionNet - Autorickshaw is incorrectly detected]{\includegraphics[width = 0.45\textwidth]{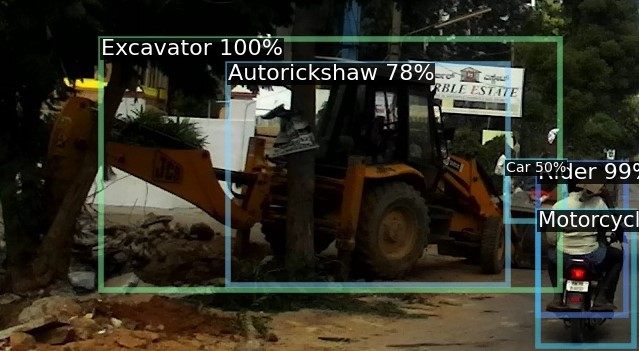}} &
\subfloat[With FusionNet - No incorrect Autorickshaw detection]{\includegraphics[width = 0.45\textwidth]{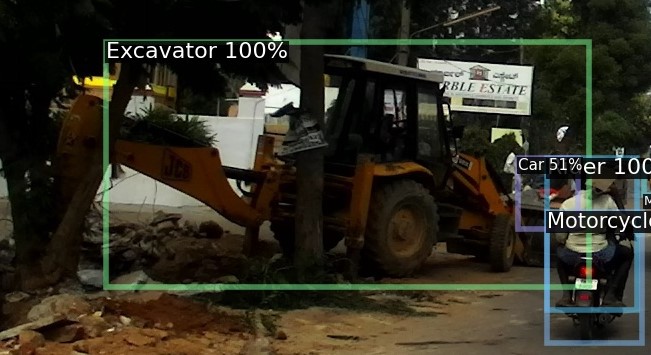}} \\  
\end{tabular}
\end{center}
\caption{Example of an erroneous prediction removed via fusion. The FusionNet is able to prevent an object from being classified as both a base class and a novel class object by handling proposals where such confusion may occur.}
\label{fig:fusion_cmp}
\end{figure*}

Figure \ref{fig:fusion_cmp} shows how the FusionNet helps reduce confusion between the base and novel classes, by handling overlapping proposal regions between the base and novel detectors. As seen in the figure, the autorickshaw class is incorrectly detected, as part of the excavator. By using the fusion approach, this region is evaluated as being part of the excavator, which gets rid of the incorrect prediction.

\subsection{COCO}
DualFusion is specifically designed for contextual datasets like IDD, as it leverages heavily on the presence of the novel classes in settings with abundant bases classes. To benchmark our DualFusion approach in a more generalized setting, we compare our network's performance to ONCE \cite{ONCE} on the COCO incremental learning task. 

We use the incremental batch of 20 classes setup as described by \cite{ONCE}. Our training methodology is identical to the training described in Section \ref{sec:df}, with the only change being the use of the COCO dataset as opposed to IDD. We treat the 20 classes in common with the PASCAL VOC \cite{pascal_voc} dataset as novel (\textit{aeroplane, bicycle, bird, boat, bottle, bus, car, cat, chair, cow, dining table, dog, horse, motorbike, person, potted plant, sheep, sofa, train} and \textit{tv/monitor}), and the remainder as base classes. We consider the 10-shot setting once again, and the training examples used are the same as \cite{fewx}. Due to computational constraints, we only trained our base Faster R-CNN on a single GPU, using the default training settings. As such, the base class Faster R-CNN detector was only able to reach a base class AP score of 19. We use the default FewX training settings to train our novel class detector. Our fusion model was trained for 10 epochs at a learning rate of 0.001. We used an IoA overlap threshold of 0.8 during the fusion training stage. The regular FewX model achieved an $AP$ score of 13.1 on the novel classes, but as stated earlier, it fails for the base cases. As such we do not include it in Table \ref{tab:dualfusion_coco}.

\begin{table}[htbp]
\centering
\begin{tabular}{l|c|c|c}
\hline
Method         & Base Classes AP & Novel Classes AP & All Classes AP \\ \hline
Fine-Tuning    & 2.8             & 0.6              & 2.3            \\ 
Feature Reweight & 3.7             & 1.5              & 3.1            \\ 
ONCE           & \textbf{17.9}            & 1.2              & \textbf{13.7}           \\ 
DualFusion (ours) & 10.8            & \textbf{9.9}              & 10.6  \\ \hline
\end{tabular}
\caption{Performance on COCO 20-Class Batch Incremental Few-shot Setting. DualFusion results in the highest novel class performance compared to the other techniques, achieving 6.6 $\times$ the performance of the closest competitor.}
\label{tab:dualfusion_coco}
\end{table}

\textbf{Results}: As seen in Table \ref{tab:dualfusion_coco} the results clearly show that DualFusion has superior novel class performance compared to ONCE, while still retaining a good degree of base class performance. The results for Finetuning, Feature Reweight and ONCE are from \cite{ONCE}. Finetuning  refers to vanilla finetuning of a CentreNet \cite{centrenet} model on novel class data, and Feature Reweight is a CentreNet implementation of \cite{feat_reweight}. Since DualFusion is designed specifically for contextual settings like driving, it is encouraging to see good performance, even in a generalized object detection challenge like COCO, achieving almost 8 times the novel class performance. Granted, our initial base class performance was only 19 AP and as such, we retain over 56\% of the base class performance in this setting, and retain roughly 76\% of FewX's novel class performance. Interestingly, we also observed that 71 of the 80 classes were detected during the Fusion Network training stage, indicating our method of generating pseudo labels for base classes is quite efficient even in generalized settings. Moreover, it enables us to train the detector with no knowledge of base classes in a truly incremental setting. These results show that our work is applicable to more than just road object detection, and is applicable to wider settings.

\section{Conclusion} \label{sec:conclusion}
In summary, through this work we present an architecture for batch incremental object detection for road object detection, that employs the simple, yet effective strategy of fusing specialized object detectors. DualFusion allows for incrementally adding the ability to detect new classes, using less data, with the only requirement being access to data from all the novel classes. This approach allows for a high degree of base class performance retention and prevention of catastrophic forgetting, while providing extremely high novel class performance. We solidify our claims by obtaining the highest base $mAP_50$ and overall $mAP_50$ scores (40.0 and 38.8 respectively), on the IDD-OS incremental task. While this technique is designed for road object detection, it is also suitable for other object detection problems where the classes of interest do not occur in isolation. This can be seen by its performance on the COCO dataset where it achieves the highest recorded novel class detection score of 9.9 AP in the 20 class few-shot batch incremental setting. 

The chief limitation of DualFusion is its reliance to permanent access to novel class data. While it requires access to the base class data only once, all the novel few-shot data must be retained. Additionally, the evaluation time grows substantially with the addition of more novel classes. Techniques to overcoming these limitations could pave the way for lifelong few-shot object detection, which in turn would greatly increase the robustness and reliability of autonomous vehicles and similar safety critical systems.

\bibliographystyle{abbrvnat}
\bibliography{bibliography/incremental_fsod}

\begin{thebibliography}{32}
\providecommand{\natexlab}[1]{#1}
\providecommand{\url}[1]{\texttt{#1}}
\expandafter\ifx\csname urlstyle\endcsname\relax
  \providecommand{\doi}[1]{doi: #1}\else
  \providecommand{\doi}{doi: \begingroup \urlstyle{rm}\Url}\fi

\bibitem[Castro et~al.(2018)Castro, Mar{\'i}n-Jim{\'e}nez, Guil, Schmid, and
  Alahari]{etoe_incremental}
F.~M. Castro, M.~J. Mar{\'i}n-Jim{\'e}nez, N.~Guil, C.~Schmid, and K.~Alahari.
\newblock End-to-end incremental learning.
\newblock In \emph{Computer Vision -- ECCV 2018}. Springer International
  Publishing, 2018.
\newblock ISBN 978-3-030-01258-8.

\bibitem[Chen et~al.(2019)Chen, Yu, and Chen]{chen_incremental_distillation}
L.~Chen, C.~Yu, and L.~Chen.
\newblock A new knowledge distillation for incremental object detection.
\newblock In \emph{2019 International Joint Conference on Neural Networks
  (IJCNN)}, 2019.
\newblock \doi{10.1109/IJCNN.2019.8851980}.

\bibitem[Everingham et~al.(2010)Everingham, Gool, Williams, Winn, and
  Zisserman]{pascal_voc}
M.~Everingham, L.~Gool, C.~K. Williams, J.~Winn, and A.~Zisserman.
\newblock The pascal visual object classes (voc) challenge.
\newblock \emph{Int. J. Comput. Vision}, 88\penalty0 (2):\penalty0 303–338,
  June 2010.
\newblock ISSN 0920-5691.
\newblock \doi{10.1007/s11263-009-0275-4}.
\newblock URL \url{https://doi.org/10.1007/s11263-009-0275-4}.

\bibitem[Fan et~al.(2020)Fan, Zhuo, Tang, and Tai]{fewx}
Q.~Fan, W.~Zhuo, C.-K. Tang, and Y.-W. Tai.
\newblock Few-shot object detection with attention-rpn and multi-relation
  detector.
\newblock In \emph{CVPR}, 2020.

\bibitem[French(1999)]{french1999catastrophic}
R.~M. French.
\newblock Catastrophic forgetting in connectionist networks.
\newblock \emph{Trends in cognitive sciences}, 3\penalty0 (4), 1999.

\bibitem[Geiger et~al.(2012)Geiger, Lenz, and Urtasun]{KITTI}
A.~Geiger, P.~Lenz, and R.~Urtasun.
\newblock Are we ready for autonomous driving? the kitti vision benchmark
  suite.
\newblock In \emph{Conference on Computer Vision and Pattern Recognition
  (CVPR)}, 2012.

\bibitem[Hao et~al.(2019)Hao, Fu, Jiang, and
  Tian]{hao_architecture_incremental}
Y.~Hao, Y.~Fu, Y.-G. Jiang, and Q.~Tian.
\newblock An end-to-end architecture for class-incremental object detection
  with knowledge distillation.
\newblock In \emph{2019 IEEE International Conference on Multimedia and Expo
  (ICME)}, 2019.
\newblock \doi{10.1109/ICME.2019.00009}.

\bibitem[Hinton et~al.(2015)Hinton, Vinyals, and Dean]{hinton_distil}
G.~Hinton, O.~Vinyals, and J.~Dean.
\newblock Distilling the knowledge in a neural network.
\newblock In \emph{NIPS Deep Learning and Representation Learning Workshop},
  2015.
\newblock URL \url{http://arxiv.org/abs/1503.02531}.

\bibitem[Kang et~al.(2019)Kang, Liu, Wang, Yu, Feng, and
  Darrell]{feat_reweight}
B.~Kang, Z.~Liu, X.~Wang, F.~Yu, J.~Feng, and T.~Darrell.
\newblock Few-shot object detection via feature reweighting.
\newblock \emph{2019 IEEE/CVF International Conference on Computer Vision
  (ICCV)}, 2019.

\bibitem[Kim et~al.(2020)Kim, Jung, and Lee]{FSOD_KT}
G.~Kim, H.-G. Jung, and S.-W. Lee.
\newblock Few-shot object detection via knowledge transfer.
\newblock In \emph{2020 IEEE International Conference on Systems, Man, and
  Cybernetics (SMC)}, pages 3564--3569, 2020.
\newblock \doi{10.1109/SMC42975.2020.9283497}.

\bibitem[Klambauer et~al.(2017)Klambauer, Unterthiner, Mayr, and
  Hochreiter]{selu}
G.~Klambauer, T.~Unterthiner, A.~Mayr, and S.~Hochreiter.
\newblock Self-normalizing neural networks.
\newblock In \emph{Advances in Neural Information Processing Systems},
  volume~30. Curran Associates, Inc., 2017.

\bibitem[Lin et~al.(2014)Lin, Maire, Belongie, Hays, Perona, Ramanan,
  Doll{\'a}r, and Zitnick]{ms-coco}
T.-Y. Lin, M.~Maire, S.~J. Belongie, J.~Hays, P.~Perona, D.~Ramanan,
  P.~Doll{\'a}r, and C.~L. Zitnick.
\newblock Microsoft coco: Common objects in context.
\newblock In \emph{ECCV}, 2014.

\bibitem[Lin et~al.(2017)Lin, Goyal, Girshick, He, and Dollár]{retinanet}
T.-Y. Lin, P.~Goyal, R.~Girshick, K.~He, and P.~Dollár.
\newblock Focal loss for dense object detection.
\newblock In \emph{2017 IEEE International Conference on Computer Vision
  (ICCV)}, 2017.
\newblock \doi{10.1109/ICCV.2017.324}.

\bibitem[Liu et~al.(2016)Liu, Anguelov, Erhan, Szegedy, Reed, Fu, and
  Berg]{SSD}
W.~Liu, D.~Anguelov, D.~Erhan, C.~Szegedy, S.~Reed, C.-Y. Fu, and A.~C. Berg.
\newblock Ssd: Single shot multibox detector.
\newblock In \emph{Computer Vision -- ECCV 2016}, pages 21--37. Springer
  International Publishing, 2016.
\newblock ISBN 978-3-319-46448-0.

\bibitem[Majee et~al.(2021)Majee, Agrawal, and Subramanian]{majee2021fewshot}
A.~Majee, K.~Agrawal, and A.~Subramanian.
\newblock Few-shot learning for road object detection.
\newblock In \emph{AAAI Workshop on Meta-Learning}, 2021.

\bibitem[McCloskey and Cohen(1989)]{mccloskey1989catastrophic}
M.~McCloskey and N.~J. Cohen.
\newblock Catastrophic interference in connectionist networks: The sequential
  learning problem.
\newblock In \emph{Psychology of learning and motivation}, volume~24, pages
  109--165. Elsevier, 1989.

\bibitem[Perez-Rua et~al.(2020)Perez-Rua, Zhu, Hospedales, and Xiang]{ONCE}
J.-M. Perez-Rua, X.~Zhu, T.~M. Hospedales, and T.~Xiang.
\newblock Incremental few-shot object detection.
\newblock In \emph{IEEE/CVF Conference on Computer Vision and Pattern
  Recognition (CVPR)}, June 2020.

\bibitem[Redmon et~al.(2016)Redmon, Divvala, Girshick, and Farhadi]{yolo}
J.~Redmon, S.~Divvala, R.~B. Girshick, and A.~Farhadi.
\newblock You only look once: Unified, real-time object detection.
\newblock \emph{2016 IEEE Conference on Computer Vision and Pattern Recognition
  (CVPR)}, 2016.

\bibitem[Ren et~al.(2015)Ren, He, Girshick, and Sun]{faster-rcnn}
S.~Ren, K.~He, R.~B. Girshick, and J.~Sun.
\newblock Faster r-cnn: Towards real-time object detection with region proposal
  networks.
\newblock \emph{IEEE Transactions on Pattern Analysis and Machine
  Intelligence}, 39, 2015.

\bibitem[Shmelkov et~al.(2017{\natexlab{a}})Shmelkov, Schmid, and
  Karteek]{Shmelkov2017IncrementalLO}
K.~Shmelkov, C.~Schmid, and A.~Karteek.
\newblock Incremental learning of object detectors without catastrophic
  forgetting.
\newblock \emph{2017 IEEE International Conference on Computer Vision (ICCV)},
  2017{\natexlab{a}}.

\bibitem[Shmelkov et~al.(2017{\natexlab{b}})Shmelkov, Schmid, and
  Karteek]{ShmelkovIncremental}
K.~Shmelkov, C.~Schmid, and A.~Karteek.
\newblock Incremental learning of object detectors without catastrophic
  forgetting.
\newblock \emph{2017 IEEE International Conference on Computer Vision (ICCV)},
  2017{\natexlab{b}}.

\bibitem[Snell et~al.(2017)Snell, Swersky, and Zemel]{proto_fewshot}
J.~Snell, K.~Swersky, and R.~Zemel.
\newblock Prototypical networks for few-shot learning.
\newblock In \emph{Advances in Neural Information Processing Systems},
  volume~30. Curran Associates, Inc., 2017.

\bibitem[Sung et~al.(2018)Sung, Yang, Zhang, Xiang, Torr, and
  Hospedales]{sung2018learning}
F.~Sung, Y.~Yang, L.~Zhang, T.~Xiang, P.~H. Torr, and T.~M. Hospedales.
\newblock Learning to compare: Relation network for few-shot learning.
\newblock In \emph{Proceedings of the IEEE Conference on Computer Vision and
  Pattern Recognition}, 2018.

\bibitem[Varma et~al.(2019)Varma, Subramanian, Namboodiri, Chandraker, and
  Jawahar]{IDD}
G.~Varma, A.~Subramanian, A.~Namboodiri, M.~Chandraker, and C.~Jawahar.
\newblock Idd: A dataset for exploring problems of autonomous navigation in
  unconstrained environments.
\newblock In \emph{2019 IEEE Winter Conference on Applications of Computer
  Vision (WACV)}, Los Alamitos, CA, USA, Jan 2019. IEEE Computer Society.
\newblock \doi{10.1109/WACV.2019.00190}.
\newblock URL
  \url{https://doi.ieeecomputersociety.org/10.1109/WACV.2019.00190}.

\bibitem[Verschae and Ruiz-del Solar(2015)]{10.3389/frobt.2015.00029}
R.~Verschae and J.~Ruiz-del Solar.
\newblock Object detection: Current and future directions.
\newblock \emph{Frontiers in Robotics and AI}, 2:\penalty0 29, 2015.
\newblock ISSN 2296-9144.
\newblock \doi{10.3389/frobt.2015.00029}.
\newblock URL
  \url{https://www.frontiersin.org/article/10.3389/frobt.2015.00029}.

\bibitem[Vinyals et~al.(2016)Vinyals, Blundell, Lillicrap, kavukcuoglu, and
  Wierstra]{fewshot_vinyals}
O.~Vinyals, C.~Blundell, T.~Lillicrap, k.~kavukcuoglu, and D.~Wierstra.
\newblock Matching networks for one shot learning.
\newblock In \emph{Advances in Neural Information Processing Systems},
  volume~29. Curran Associates, Inc., 2016.

\bibitem[Wang et~al.(2020{\natexlab{a}})Wang, Huang, Darrell, Gonzalez, and
  Yu]{fsdet}
X.~Wang, T.~E. Huang, T.~Darrell, J.~E. Gonzalez, and F.~Yu.
\newblock Frustratingly simple few-shot object detection.
\newblock July 2020{\natexlab{a}}.

\bibitem[Wang et~al.(2020{\natexlab{b}})Wang, Yao, Kwok, and
  Ni]{few_shot_survey}
Y.~Wang, Q.~Yao, J.~T. Kwok, and L.~M. Ni.
\newblock Generalizing from a few examples: A survey on few-shot learning.
\newblock \emph{ACM Comput. Surv.}, 53\penalty0 (3), June 2020{\natexlab{b}}.
\newblock ISSN 0360-0300.
\newblock \doi{10.1145/3386252}.
\newblock URL \url{https://doi.org/10.1145/3386252}.

\bibitem[Wang et~al.(2019)Wang, Ramanan, and Hebert]{metadet}
Y.-X. Wang, D.~Ramanan, and M.~Hebert.
\newblock Meta-learning to detect rare objects.
\newblock In \emph{Proceedings of the IEEE/CVF International Conference on
  Computer Vision (ICCV)}, October 2019.

\bibitem[Wu et~al.(2020)Wu, Sahoo, and Hoi]{meta_rcnn}
X.~Wu, D.~Sahoo, and S.~Hoi.
\newblock Meta-rcnn: Meta learning for few-shot object detection.
\newblock In \emph{Proceedings of the 28th ACM International Conference on
  Multimedia}, MM '20, page 1679–1687, New York, NY, USA, 2020. Association
  for Computing Machinery.
\newblock ISBN 9781450379885.
\newblock \doi{10.1145/3394171.3413832}.
\newblock URL \url{https://doi.org/10.1145/3394171.3413832}.

\bibitem[Yu et~al.(2020)Yu, Chen, Wang, Xian, Chen, Liu, Madhavan, and
  Darrell]{bdd100k}
F.~Yu, H.~Chen, X.~Wang, W.~Xian, Y.~Chen, F.~Liu, V.~Madhavan, and T.~Darrell.
\newblock Bdd100k: A diverse driving dataset for heterogeneous multitask
  learning.
\newblock In \emph{IEEE/CVF Conference on Computer Vision and Pattern
  Recognition (CVPR)}, June 2020.

\bibitem[Zhou et~al.(2019)Zhou, Wang, and Krähenbühl]{centrenet}
X.~Zhou, D.~Wang, and P.~Krähenbühl.
\newblock Objects as points, 2019.

\end{thebibliography}

\end{document}